\documentclass{article} 
\usepackage{iclr2027_conference,times}

\usepackage[utf8]{inputenc} 
\usepackage[T1]{fontenc}    
\usepackage{amsmath}
\usepackage{amsfonts}       
\usepackage{booktabs}       
\usepackage{nicefrac}       
\usepackage{microtype}      
\usepackage{xcolor}         
\usepackage{graphicx}
\graphicspath{{figures/}} 

\usepackage{amsmath,amsfonts,bm}

\def\eqref#1{equation~\ref{#1}}

\def\1{\bm{1}}

\DeclareMathAlphabet{\mathsfit}{\encodingdefault}{\sfdefault}{m}{sl}
\SetMathAlphabet{\mathsfit}{bold}{\encodingdefault}{\sfdefault}{bx}{n}

\usepackage{hyperref}
\usepackage{url}

\title{When Data Imbalance Helps: Robust Generalization Through Shortcut Saturation}

\author{%
  Cheng-Ting Chou\thanks{Equal contribution.} \\
  University of California, Los Angeles \\
  \texttt{ctchou3@cs.ucla.edu} \\
  \And
  Duc Binh Hoang\footnotemark[1] \\
  Purdue University \\
  \texttt{hoang112@purdue.edu} \\
}

\newcommand{\us}{\_\allowbreak}
\newcommand{\tsk}[1]{\texttt{#1}}

\iclrfinalcopy 
\begin{document}

\maketitle
\lhead{}
\raggedbottom

\begin{abstract}
We study robust generalization under spurious correlations: tasks where a shortcut feature is correlated with the true label in training, but anti-correlated in an adversarial held-out split. Varying the spurious ratio $r$ (the fraction of training examples where shortcut = true label) and model capacity, we find a counterintuitive result: data imbalance promotes generalization in sufficiently capable models. On a synthetic task where the true label is sum parity of an integer sequence and the shortcut is the parity of the maximum-valued element, a 2-layer, 2-head transformer generalized (reached $100\%$ adversarial accuracy) in 0\% of seeds at $r{=}0.50$ but 77\% of seeds at $r{=}0.90$. The effect is absent in 1-layer models, where imbalance instead traps the model on the shortcut. Through mechanistic analysis---gradient conflict dynamics, circuit evolution, and query-key/output-value (QK/OV) circuit ablations---we characterize a mechanistic pathway consistent with imbalance promoting generalization.
\end{abstract}

\section{Introduction}

Spurious correlations are one of the central failure modes of learned models \citep{geirhos2020shortcut, shah2020pitfalls}. When a feature is highly predictive in training but unreliable at test time, models learn to exploit it as a shortcut, and the standard prescriptions correct for this by intervening on the group structure of the data---minimizing worst-group risk \citep{sagawa2020distributionally} or subsampling the majority group so that shortcut-consistent examples no longer dominate gradient descent \citep{sagawa2020overparameterization}.

We identify a regime in which the balancing half of that advice is counterproductive. Subsampling the majority group drives the shortcut-consistent fraction toward the random-chance baseline, and for models above a capacity threshold our sweeps isolate that baseline as the \emph{worst} available setting: six of the eight two-layer configurations we test generalize in 0\% of seeds there and the best reaches 10\%, while ratios far from the baseline reach 77--93\% on three of the four tasks. We do not implement group-level reweighting itself, so this speaks to the balancing target such methods move toward, not to their full machinery.

In a controlled synthetic setting where models must predict sum parity of an integer sequence but can exploit the parity of the maximum-valued element as a shortcut (the Max-Parity-Sum-Parity task, defined in Section~\ref{sec:setup}), we find that increasing the spurious ratio $r$---the fraction of training examples where shortcut equals true label---from 0.50 to 0.90 substantially increases the probability of robust generalization, but only in models with sufficient capacity. Two-layer, two-head transformers generalize in 0\% of seeds at $r{=}0.50$ and 77\% of seeds at $r{=}0.90$; single-layer models show the opposite trend, with higher imbalance trapping them on the shortcut.

We hypothesize that the mechanism underlying this reversal involves \emph{shortcut saturation}---the regime in which shortcut-consistent examples achieve near-zero loss, causing their gradients to vanish. When imbalance is high, the shortcut circuit rapidly achieves near-perfect accuracy on the majority of training examples, bringing them into this saturated regime. The anti-shortcut minority, consistently misclassified, continues to produce large gradients---amplified, as we argue in Section~\ref{sec:discussion}, by at least a factor of $r/(1{-}r)$, or 9:1 at $r{=}0.90$. In capable models, this amplified adversarial gradient appears to support a structural reorganization of the attention circuit. At balanced ratios, no such saturation appears to occur: the two gradient sources remain persistently opposed and neither side gains sufficient gradient momentum. These results suggest that shortcut saturation may function as a precondition for generalization rather than an obstacle to it.

Our contributions are:
\begin{enumerate}
  \item \textbf{Imbalance $\times$ capacity interaction.} Increasing the spurious ratio from 0.5 to 0.9 raises the two-layer, two-head generalization rate from 0\% to 77\% on both binary-label tasks, while one-layer models do not benefit: on the magnitude-shortcut task their rate falls over the same sweep (33\% to 0\% at one head, 10\% to 3\% at two), and on the position-shortcut task it peaks at the intermediate ratio and returns to 3--7\% at $r{=}0.9$. This places a capacity threshold between one and two transformer layers on both binary-label tasks. The threshold is task-dependent: it sits below two layers on the easier n11 variant, and the ordering inverts on the ternary magnitude-shortcut task (Appendix~\ref{app:mod3_analysis}).
  \item \textbf{Mechanistic pathway.} Gradient conflict analysis, circuit evolution tracking, and QK/OV ablations characterize a pathway consistent with shortcut saturation amplifying adversarial gradients and supporting structural circuit reorganization in capable models.
  \item \textbf{Unifying principle.} Across four tasks (two shortcut types $\times$ two label bases), the effect tracks the deviation of $r$ from the random-chance baseline rather than the absolute value of $r$: for two-layer models the null ratio is at or below every other ratio tested on all four tasks. On the two ternary tasks, whose sweeps straddle the baseline, this holds in both directions: anti-shortcut-biased ratios are associated with generalization as strongly as shortcut-biased ones.
\end{enumerate}

\section{Background and Related Work}

\paragraph{Grokking and Delayed Generalization.}
The phenomenon of grokking---a sharp phase transition in test accuracy long after training loss has plateaued---was identified by \citet{power2022grokking} on modular arithmetic tasks. Subsequent work has characterized what drives these transitions from two directions: mechanistic analysis of the circuit that forms during the transition \citep{nanda2023progress}, and progress measures showing that learning advances continuously while loss and error remain flat \citep{barak2022hidden}. Our setting is related but distinct: we observe a continuum of outcomes (some runs generalize quickly, some slowly, some not at all) without assuming or requiring a sharp phase transition.

\paragraph{Shortcut Learning and Simplicity Bias.}
Deep neural networks are known to exhibit a \textit{simplicity bias}, a tendency to rely on the simplest available features that decrease training error \citep{shah2020pitfalls, teney2025simplicity}. This bias leads to \textit{shortcut learning} in the presence of spurious correlations---where models exploit easy-to-represent but non-causal patterns \citep{geirhos2020shortcut}. Recent work suggests that this bias is a fundamental characteristic of deep non-linear architectures, and that which features a model uses depends jointly on a feature's \textit{predictivity}---how reliably it indicates the training labels---and its \textit{availability}---how easily it can be extracted from the input; these findings are consistent with a theoretical account based on Neural Tangent Kernels (NTK) \citep{hermann2024foundations}.

\paragraph{The Paradox of Data Imbalance.}
Standard prescriptions for mitigating shortcuts intervene on the training signal so that no single feature or group dominates it: worst-group risk minimization with strong regularization \citep{sagawa2020distributionally}, subsampling the majority group \citep{sagawa2020overparameterization}, difficulty-based loss reweighting \citep{sinha2020consistency}, and surgery on conflicting per-domain gradients \citep{mansilla2021domain}. We instead identify a regime where imbalance acts as a catalyst. Related work on \textit{spurious memorization} localizes a model's fit to atypical minority-group examples in a small subset of neurons \citep{you2025uncovering}; our account is distinct and, to our knowledge, untested there: once the shortcut saturates, the signal from shortcut-consistent examples diminishes, amplifying the relative gradient contribution of the anti-shortcut minority enough to overcome the simplicity bias in capable models.

\paragraph{Mechanistic Interpretability of Circuits.}
Our approach builds on mechanistic interpretability techniques designed to identify internal circuits within Transformers \citep{nanda2023progress}. While prior studies have focused on prefix-matching induction heads \citep{olsson2022context} or the Fourier-based modular arithmetic circuit later named the ``Clock'' algorithm \citep{nanda2023progress, zhong2023clock}, we focus on the transition between magnitude-sorting and parity-encoding circuits.

\section{Setup}
\label{sec:setup}
\subsection{Tasks}

We study four tasks formed by the cross-product of two true labels and two shortcut features, all using sequences of 5 integers drawn i.i.d.\ from $[0,20)$. The \textbf{true labels} are \emph{sum parity} (sum mod~2) and \emph{sum mod~3}; the \textbf{shortcut features} are the \emph{first element} or the \emph{max-valued element}, each taken modulo the same base as the label. The four combinations give the task names used throughout: Max-Parity-Sum-Parity (MPSP) and First-Equal-Sum-Parity (FESP) for the binary labels, and First-Equal-Sum-Mod3 (FE-Mod3) and Max-Mod3-Sum-Mod3 (MM-Mod3) for the ternary ones. An example is \emph{shortcut-consistent} when the shortcut equals the true label, and \emph{anti-shortcut} otherwise. The \textbf{spurious ratio} $r$ is the fraction of shortcut-consistent training examples. Because the random-chance baseline differs by label type---$\nicefrac{1}{2}$ for binary (mod~2) tasks and $\nicefrac{1}{3}$ for ternary (mod~3) tasks---we use task-appropriate ratio sweeps: $r \in \{0.5, 0.7, 0.9\}$ for parity tasks and $r \in \{0.15, 0.33, 0.5\}$ for mod~3 tasks. Attending to either the max or the first element is mechanistically cheaper than aggregating all elements, creating a capacity asymmetry that is exploitable by label imbalance.

\subsection{Dataset Splits}

Each experiment uses three fixed splits: a \textbf{train} set of 16{,}384 examples at spurious ratio $r$; a \textbf{val} set of 2{,}048 examples at ratio 1.00 (all shortcut-consistent), measuring shortcut learning; and an \textbf{adv} (adversarial) set of 2{,}048 examples at ratio 0.00 (all anti-shortcut), measuring true-rule generalization.

\paragraph{Generalization definition.}
A run \emph{generalizes} if it achieves $100\%$ adversarial accuracy---all 2{,}048 adversarial examples correct---at any logged epoch, accuracy being logged every 100 epochs. The \emph{generalization epoch} is the first such epoch. The \emph{generalization rate} per configuration is the fraction of seeds where the generalization epoch is finite. No assumption is made about whether the improvement is sudden or gradual.

\subsection{Models and Training}

\textbf{Architecture variants} are single-layer and two-layer transformers with $n_\text{heads} \in \{1, 2\}$, $d_\text{model}{=}64$, and $d_\text{ff}{=}128$. \textbf{Training} runs for 15{,}000 epochs with AdamW ($\text{lr}{=}10^{-3}$, weight decay $\in \{0.1, 0.4\}$, batch size 4{,}096). Each configuration is run with 30 seeds for the two primary binary-label tasks (MPSP and FESP) and the two mod~3 tasks (FE-Mod3 and MM-Mod3), and 10 seeds for the remaining ablation variants (n11; Appendix~\ref{app:n11}); compute details are in Appendix~\ref{app:compute}. Shortcut-consistent and adversarial accuracy are logged every 100 epochs; checkpoints are saved at every 10\% absolute change in adversarial accuracy.

\section{Behavioral Results: Imbalance and Capacity Interaction}

This section covers the two binary-label tasks (max-element and first-element shortcuts); ternary mod~3 tasks are summarized at the end of Section~\ref{sec:cross-shortcut} and detailed in Appendix~\ref{app:mod3_analysis}.

\subsection{Generalization Rate Increases With Ratio for 2-Layer Models}
\label{sec:gen-rate}
\begin{table}[b]
  \caption{Generalization rate by configuration (weight decay = 0.1). Bold rows highlight the 2-layer models where the imbalance $\times$ capacity interaction is most pronounced.}
  \label{tab:main}
  \centering
  \begin{tabular}{ll@{\hskip 1.5em}ll}
    \toprule
    \multicolumn{2}{c}{1-layer models} & \multicolumn{2}{c}{2-layer models} \\
    \cmidrule(r){1-2}\cmidrule(l){3-4}
    Config & Gen.\ Rate & Config & Gen.\ Rate \\
    \midrule
    1H, $r{=}0.5$ & 10/30 (33\%) & \textbf{1H, $r{=}0.5$} & \textbf{3/30 (10\%)} \\
    1H, $r{=}0.7$ & 2/30 (7\%)   & \textbf{1H, $r{=}0.7$} & \textbf{8/30 (27\%)} \\
    1H, $r{=}0.9$ & 0/30 (0\%)   & \textbf{1H, $r{=}0.9$} & \textbf{16/30 (53\%)} \\
    2H, $r{=}0.5$ & 3/30 (10\%)  & \textbf{2H, $r{=}0.5$} & \textbf{0/30 (0\%)} \\
    2H, $r{=}0.7$ & 3/30 (10\%)  & \textbf{2H, $r{=}0.7$} & \textbf{13/30 (43\%)} \\
    2H, $r{=}0.9$ & 1/30 (3\%)   & \textbf{2H, $r{=}0.9$} & \textbf{23/30 (77\%)} \\
    \bottomrule
  \end{tabular}
\end{table}

Table~\ref{tab:main} reveals the key interaction: the 2L, 2H model's generalization rate rises from 0\% to 43\% to 77\% as $r$ increases from 0.5 to 0.7 to 0.9.

For 1-layer models, the pattern inverts: the 1L, 1H model generalizes in 33\% of seeds at balanced training, but this collapses to 7\% and then 0\% as the ratio increases---higher imbalance actively traps training on the shortcut. This is consistent with a capacity threshold below which imbalance is harmful and above which it is beneficial. For this task, the threshold lies between one and two transformer layers.

\subsection{Weight Decay Causes Transient Generalization}
\label{sec:wd-transient}

Increasing weight decay from 0.1 to 0.4 leaves the overall generalization rate largely intact---the 2L, 2H model still has 21/30 seeds reach 100\% adversarial accuracy at some point during training at $r{=}0.9$---but per-seed inspection reveals that 15 of those 21 seeds hit it only transiently before regressing to near-shortcut performance; only 6 consolidated into stable generalizing solutions. Higher weight decay thus enables models to discover the generalizing basin but prevents them from staying in it. Per-configuration results are in Appendix~\ref{app:wd04}.

\subsection{Shortcut-Trapping in 1-Layer Models}

One-layer models at $r \geq 0.7$ present a clear picture of shortcut reliance: they reach near-perfect accuracy on shortcut-consistent examples (0.90--1.00 mean per configuration) while adversarial accuracy stays near zero (0.01--0.20) throughout the entire 15{,}000-epoch training run. Note that near zero is the signature of shortcut reliance, not chance: because the adversarial split is constructed so the shortcut contradicts the label, a model that has fully absorbed the shortcut is systematically wrong rather than random. This is not a transient state or a convergence artifact---it is a stable configuration from which training does not recover within 15,000 epochs.

\subsection{Cross-Shortcut Generalization: Position vs.\ Magnitude}
\label{sec:cross-shortcut}

The behavioral effect replicates across a structurally distinct shortcut type. In the \tsk{first\us{}equal\us{}sum\us{}parity} (FESP) task, the shortcut is purely positional: the parity of the first sequence element is correlated with sum parity in training. Where the magnitude shortcut requires the model to rank tokens by magnitude, the position shortcut requires only reading the value at a fixed position.

\begin{table}[htbp]
  \caption{Generalization rates for 2-layer models (wd=0.1): MPSP (magnitude shortcut) vs.\ FESP (position shortcut). Full results including 1-layer models are in Appendix~\ref{app:fesp_analysis}.}
  \label{tab:cross-shortcut}
  \centering
  \small
  \begin{tabular}{ll@{\hskip 1em}cc}
    \toprule
    Config & $r$ & MPSP (mag.\ shortcut) & FESP (pos.\ shortcut) \\
    \midrule
    2L, 1H & 0.5 & 3/30 (10\%)  & 0/30 (0\%) \\
    2L, 1H & 0.7 & 8/30 (27\%)  & 16/30 (53\%) \\
    2L, 1H & 0.9 & 16/30 (53\%) & 20/30 (67\%) \\
    2L, 2H & 0.5 & 0/30 (0\%)   & 0/30 (0\%) \\
    2L, 2H & 0.7 & 13/30 (43\%) & 20/30 (67\%) \\
    \textbf{2L, 2H} & \textbf{0.9} & \textbf{23/30 (77\%)} & \textbf{23/30 (77\%)} \\
    \bottomrule
  \end{tabular}
\end{table}

The qualitative pattern is preserved across both shortcut types (Table~\ref{tab:cross-shortcut}): 2-layer models benefit from increasing imbalance, with generalization rising as $r$ increases. The two shortcut types reach comparable peak generalization---the best position-shortcut configuration reaches 77\% at $r{=}0.9$, matching the 77\% of the magnitude-shortcut task at the same setting. At intermediate ratios the position shortcut generalizes somewhat more readily (e.g., 2L, 1H at $r{=}0.7$: 53\% FESP vs.\ 27\% MPSP). This indicates that a purely positional shortcut is no more resistant to displacement than a magnitude-based one: both are dislodged by the amplified adversarial gradient once the shortcut saturates and the model has sufficient capacity.

\paragraph{Ternary label replication.}
The pattern extends to ternary labels. In the \tsk{first\us{}equal\us{}sum\us{}mod3} (FE-Mod3) task, the true label is sum mod~3 (three classes) and the shortcut is the first element mod~3. The random-chance shortcut correlation for a ternary problem is $\nicefrac{1}{3}$, not $\nicefrac{1}{2}$, so the ratio sweep is $r \in \{0.15, 0.33, 0.50\}$---with $r{=}0.33$ as the null point (no shortcut information) and $r{=}0.15$ and $r{=}0.50$ as imbalanced ratios in opposite directions. The 2L, 2H model achieves 93\% generalization at $r{=}0.15$ and 90\% at $r{=}0.50$, but 0\% at the null ratio ($r{=}0.33$), with 1-layer models showing substantially lower rates throughout. These results suggest that the relevant quantity is not the absolute value of $r$ but its deviation from the random-chance baseline: any imbalance---whether the shortcut is over- or under-represented---appears to create a gradient asymmetry associated with the effect, while shortcut neutrality (random chance) eliminates it entirely. Full mod~3 results are in Appendix~\ref{app:mod3_analysis}.

\begin{figure}[t]
  \centering
  \includegraphics[width=0.85\linewidth]{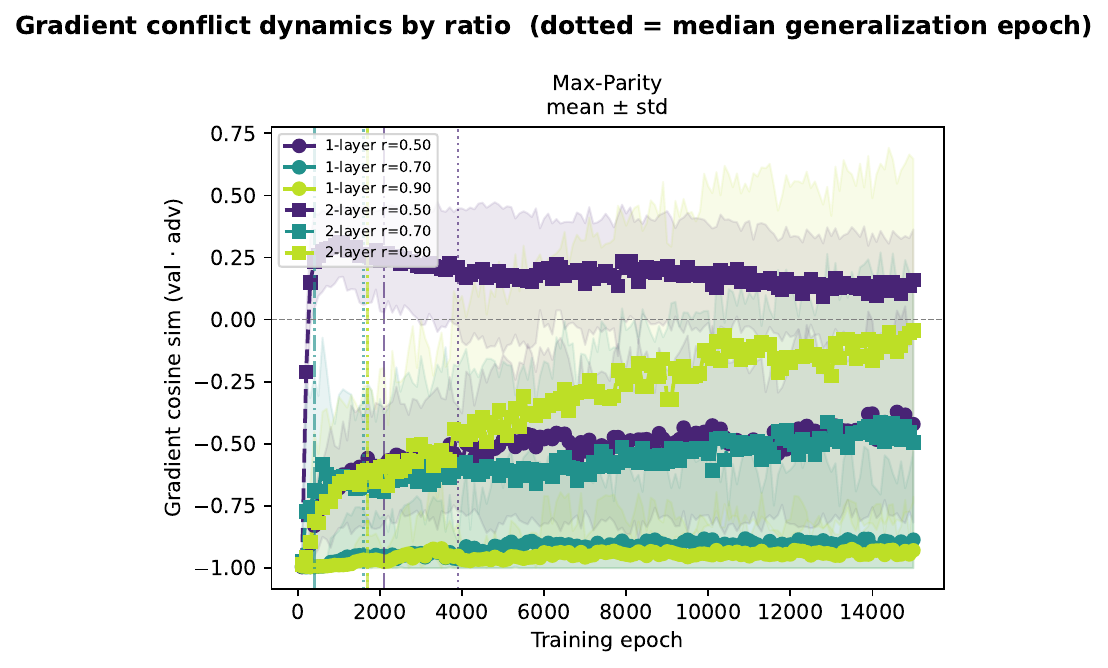}
  \caption{Gradient cosine similarity between shortcut-consistent and anti-shortcut gradient vectors over training epochs. Values near $-1$ indicate the two subsets are pulling parameters in opposite directions; values near $0$ indicate the two gradients are orthogonal, i.e.\ no longer in direct opposition. Each curve is one depth $\times$ ratio group at weight decay $0.1$, pooling both head counts (mean $\pm$ std over 60 runs; the shaded band is clipped to the attainable range $[-1,1]$). Dotted vertical lines mark the median generalization epoch of each group. Conflict persists throughout training in 1-layer models at $r \geq 0.7$ (both curves end near $-0.9$), while 2-layer models move toward zero, reaching $-0.04$ at $r{=}0.9$. Proximity to zero does not by itself indicate generalization: the 2-layer $r{=}0.5$ curve ends highest of all six ($0.16$) despite a 0\% generalization rate at 2H (Table~\ref{tab:gradient}).}
  \label{fig:gradient_dynamics}
\end{figure}

\section{Mechanistic Analysis}

The mechanistic analyses in this section are conducted on the magnitude-shortcut task (max-element parity as shortcut, defined in Section~\ref{sec:setup}).

The behavioral results establish that imbalance and capacity interact to produce different generalization outcomes. The analyses that follow characterize a mechanistic pathway consistent with the behavioral findings: imbalance is associated with amplified adversarial gradient signal (Section~\ref{sec:gradient}), which in capable models appears to support a structural reorganization of the first attention head from a shortcut circuit to a robust one (Section~\ref{sec:circuit}), a transition characterized by the circuit's query-key geometry and output-value encoding (Section~\ref{sec:qkov}). We note that these analyses establish correlations and are consistent with the proposed pathway; direct causal interventions remain future work.

\subsection{Gradient Conflict Dynamics}
\label{sec:gradient}

To understand how the two subsets interact during training, we compute the cosine similarity between the gradient obtained from shortcut-consistent examples and the gradient from anti-shortcut examples. When this quantity is strongly negative, the two subsets are pulling the model's parameters in opposite directions; when it approaches zero, their gradient signals are orthogonal rather than opposed. We also track the ratio of anti-shortcut to shortcut gradient norms, which measures the relative magnitude of the two subsets' training signals.

Gradient conflict appears early in every configuration, whether or not the model generalizes. Shortcut-trapped one-layer models at $r \geq 0.7$ hold cosine similarity near $-0.9$ throughout training---the conflict never resolves (Figure~\ref{fig:gradient_dynamics}). The means in Table~\ref{tab:gradient} carry standard deviations comparable to the mean, so we read them per seed. Conditioned on outcome, final cosine similarity separates the two groups in every configuration where both occur---generalizing versus non-generalizing seeds end at $-0.138$ versus $-0.617$, $-0.587$ versus $-0.979$, $-0.087$ versus $-0.813$ and $0.097$ versus $-0.649$ (1L,~1H at $r{=}0.5$ and $0.7$; 2L,~2H at $r{=}0.7$ and $0.9$)---with non-generalizing seeds tightly clustered ($\sigma \leq 0.16$). One configuration breaks the pattern: 2L,~2H at $r{=}0.5$ resolves conflict in 28/30 seeds ($0.122 \pm 0.128$) yet generalizes in none---it is one of the two configurations that resolve conflict without any gradient amplification, and both are two-layer models at $r{=}0.5$: 2L,~2H (ratio $0.987$, 0/30 seeds generalize) and 2L,~1H (ratio $1.043$, $0.197 \pm 0.255$ final cosine similarity, 3/30). These are the two weakest two-layer cells in Table~\ref{tab:main}. Resolving conflict in the absence of amplification therefore accompanies the worst outcomes rather than the best: conflict resolution accompanies generalization without producing it on its own.

The gradient norm ratio completes the picture. At $r{=}0.9$, anti-shortcut examples produce roughly 9--17$\times$ more gradient signal than shortcut-consistent ones (Table~\ref{tab:gradient}). This amplification arises because the saturated shortcut circuit classifies shortcut-consistent examples correctly with high confidence---their losses collapse toward zero and their gradients vanish---while the misclassified anti-shortcut minority keeps producing large gradients. In capable models this sustained minority signal is associated with eventual circuit reorganization.

\begin{table}[t]
  \caption{Gradient conflict dynamics by configuration (weight decay = 0.1). ``Gradient ratio'' is the anti-shortcut to shortcut gradient norm ratio at end of training. Entries without a standard deviation have zero variance across seeds at the reported precision.}
  \label{tab:gradient}
  \centering
  \small
  \begin{tabular}{lllll}
    \toprule
    Config & Peak conflict epoch & Min cosine sim & Final cosine sim & Gradient ratio \\
    \midrule
    1L, 1H, $r{=}0.5$ & $1070 \pm 3149$ & $-0.998 \pm 0.002$ & $-0.458 \pm 0.422$ & $0.996 \pm 0.522$ \\
    1L, 1H, $r{=}0.7$ & $287 \pm 168$ & $-0.999$ & $-0.953 \pm 0.137$ & $2.366 \pm 0.429$ \\
    1L, 1H, $r{=}0.9$ & $400 \pm 236$ & $-0.999 \pm 0.001$ & $-0.978 \pm 0.009$ & $\mathbf{8.843 \pm 0.691}$ \\
    2L, 2H, $r{=}0.5$ & 100 & $-0.978 \pm 0.022$ & $0.122 \pm 0.128$ & $0.987 \pm 0.202$ \\
    2L, 2H, $r{=}0.7$ & $1257 \pm 2305$ & $-0.991 \pm 0.023$ & $-0.498 \pm 0.587$ & $2.568 \pm 1.326$ \\
    2L, 2H, $r{=}0.9$ & $1917 \pm 3987$ & $-0.993 \pm 0.006$ & $\mathbf{-0.077 \pm 0.697}$ & $\mathbf{16.824 \pm 12.664}$ \\
    \bottomrule
  \end{tabular}
\end{table}

\subsection{Circuit Evolution}
\label{sec:circuit}

We identify a head as having developed a \emph{robust circuit} at checkpoint $t$ via activation ablation and cross-split patching. A head is labeled robust (\emph{shared}) when ablating it reduces accuracy on both the shortcut-consistent and adversarial splits (positive contribution to both) and cross-split activation patching transfers performance in both directions; for the first-formation event reported here we additionally require the mean cross-split patching score to exceed 0.5. Full criteria, including where that threshold does and does not apply, are given in Appendix~\ref{app:methods}. This gives us a per-head, per-checkpoint label that tracks when each head transitions from shortcut-implementing to true-rule-implementing circuitry; we report the earliest epoch at which any head across all layers meets this criterion.

\begin{table}[b]
  \caption{Epoch of first robust head formation, wd=0.1. Each mean $\pm$ std is taken over only the seeds that ever form a robust head; the trailing fraction gives that count out of 30, and cells with small counts are conditional on a rare event. One-layer, one-head models form a robust head rarely and very late ($12850 \pm 919$, 2/30 seeds at $r{=}0.7$) or never ($r{=}0.9$). The two-layer $r{=}0.9$ comparison is the one contrast between cells with comparable counts: doubling from 1H (27/30) to 2H (30/30) accelerates first robust head formation by $\sim$3$\times$ ($3226 \to 960$ epochs). One interpretation is that shortcut specialization on one head concentrates the growing adversarial gradient on the other. Formation is not sufficient for generalization: 1L,~2H at $r{=}0.9$ forms a robust head in 9/30 seeds but generalizes in 1/30.}
  \label{tab:circuit}
  \centering
  \small
  \begin{tabular}{ll@{\hskip 1.5em}ll}
    \toprule
    \multicolumn{2}{c}{1-layer models} & \multicolumn{2}{c}{2-layer models} \\
    \cmidrule(r){1-2}\cmidrule(l){3-4}
    Config & First robust head ~{\scriptsize $n$/30} & Config & First robust head ~{\scriptsize $n$/30} \\
    \midrule
    1H, $r{=}0.5$ & $4990 \pm 4060$ ~{\scriptsize 10/30} & 1H, $r{=}0.5$ & $4100 \pm 4068$ ~{\scriptsize 9/30} \\
    1H, $r{=}0.7$ & $12850 \pm 919$ ~{\scriptsize 2/30}  & 1H, $r{=}0.7$ & $2988 \pm 3969$ ~{\scriptsize 8/30} \\
    1H, $r{=}0.9$ & --- (never) ~{\scriptsize 0/30}      & 1H, $r{=}0.9$ & $3226 \pm 3605$ ~{\scriptsize 27/30} \\
    2H, $r{=}0.5$ & $4775 \pm 6186$ ~{\scriptsize 4/30}  & 2H, $r{=}0.5$ & $4291 \pm 4154$ ~{\scriptsize 23/30} \\
    2H, $r{=}0.7$ & $567 \pm 321$ ~{\scriptsize 3/30}    & 2H, $r{=}0.7$ & $1464 \pm 3384$ ~{\scriptsize 14/30} \\
    2H, $r{=}0.9$ & $7567 \pm 4893$ ~{\scriptsize 9/30}  & 2H, $r{=}0.9$ & $\mathbf{960 \pm 1331}$ ~{\scriptsize 30/30} \\
    \bottomrule
  \end{tabular}
\end{table}

One-layer, one-head models at $r \geq 0.7$ almost never satisfy the robust circuit criterion (2/30 and 0/30 seeds), remaining in the magnitude-sorting regime throughout training, and the two seeds that do form a robust head reach it very late (Table~\ref{tab:circuit}). One-layer, two-head models form one more often at $r{=}0.9$ (9/30) but still generalize in only 1/30 seeds, so formation alone is not sufficient.

Two-layer models at $r{=}0.9$, by contrast, undergo a structural transition whose speed depends critically on head count. In the 2L, 1H model, the robust circuit forms across both layers, with the first robust head appearing at epoch $3226 \pm 3605$ on average. In the 2L, 2H model, one head takes the shortcut early; this division of labor appears to concentrate the adversarial gradient on the second head, which develops the robust circuit far sooner (epoch $960 \pm 1331$, roughly $3\times$ faster). Crucially, at the generalization epoch itself, 96--100\% of all attention heads in the two-layer models satisfy the robust circuit criterion---regardless of when that epoch falls---suggesting a threshold effect: generalization occurs once sufficient circuit reorganization has accumulated, not at a fixed epoch. Post-generalization, some seeds regress: final-checkpoint robust fractions (0.63--0.67) are substantially lower than at-generalization fractions (0.97--1.00), consistent with the same mechanism underlying the transient generalization under higher weight decay (Section~\ref{sec:wd-transient}).

\subsection{QK/OV Circuit Fingerprint of Shortcut vs.\ Robust Models}
\label{sec:qkov}

The circuit evolution analysis identifies when the transition happens; the QK/OV fingerprint characterizes what structurally changes. In a shortcut-implementing head, the query-key circuit is consistent with magnitude ordering: higher-valued elements receive higher attention weights, and the head appears to preferentially select the maximum-valued token. In a robust head, the output-value (OV) circuit is instead consistent with parity encoding, projecting attended representations onto a parity direction rather than a magnitude direction.

\begin{table}[htbp]
  \caption{QK/OV circuit fingerprint across training. QK Spearman $\rho$ (peak) is the maximum across all checkpoints and heads; QK Spearman $\rho$ (final) is the value at the last checkpoint. OV parity Spearman (at QK peak) is measured at the epoch when QK Spearman is highest. OV parity Spearman (at generalization) is measured at the first epoch at which adversarial accuracy reaches 100\% for generalizing models; for shortcut-trapped models that never generalize, the final checkpoint value is reported instead. Because the parity target is binary (10 even and 10 odd token values), tied ranks cap the maximum attainable OV parity Spearman at $0.867$; values near $0.86$ are therefore at ceiling.}
  \label{tab:qkov}
  \centering
  \small
  \begin{tabular}{lcccc}
    \toprule
    & \multicolumn{2}{c}{Shortcut circuit (QK $\rho$)} & \multicolumn{2}{c}{Robust indicator (OV parity Spearman)} \\
    Config & Peak & Final & At QK peak & At gen. \\
    \midrule
    1L, 1H, $r{=}0.5$ \\ \small(mixed) & $0.213 \pm 0.104$ & $0.111 \pm 0.101$ & $0.033 \pm 0.361$ & $-0.062 \pm 0.568$ \\
    1L, 1H, $r{=}0.9$ \\ \small(shortcut-trapped) & $\mathbf{0.987 \pm 0.005}$ & $\mathbf{0.971 \pm 0.014}$ & $0.741 \pm 0.151$ & $0.628 \pm 0.232$ \\
    2L, 2H, $r{=}0.5$ \\ \small(balanced-trapped) & $0.295 \pm 0.090$ & $0.217 \pm 0.090$ & $0.290 \pm 0.116$ & $0.284 \pm 0.116$ \\
    2L, 2H, $r{=}0.9$, wd=0.1 \\ \small(generalized) & $\mathbf{0.979 \pm 0.011}$ & $0.682 \pm 0.225$ & $0.800 \pm 0.072$ & $\mathbf{0.864 \pm 0.010}$ \\
    \bottomrule
  \end{tabular}
\end{table}

All models with $r \geq 0.7$ develop a strong shortcut circuit early (Table~\ref{tab:qkov}): QK Spearman $\rho$ peaks between 0.95 and 0.99 during training regardless of whether the model ultimately generalizes. The critical divergence occurs afterward. In the shortcut-trapped 1L, 1H model at $r{=}0.9$, QK $\rho$ stagnates near its peak at the final checkpoint ($0.971$), and OV parity Spearman at the final checkpoint remains below its at-peak value ($0.628$ vs.\ $0.741$ at peak)---consistent with continued reliance on the shortcut representation while parity-aligned structure diminishes. In the generalizing 2L, 2H model, QK $\rho$ drops sharply from its peak of 0.979 to 0.682 at the final checkpoint, while OV parity Spearman rises from 0.800 at the QK peak to $0.864 \pm 0.010$ at the generalization epoch---effectively at the $0.867$ ceiling. The shortcut circuit forms at similar strength in all high-$r$ models; what distinguishes generalizing models is that it is subsequently displaced as the robust parity circuit matures.

The 2L, 2H model at $r{=}0.5$ exposes a qualitatively different failure mode: QK Spearman peaks at only $0.295 \pm 0.090$---far below the ${\sim}0.98$ reached at $r{=}0.9$ for the same architecture---indicating that no strong shortcut circuit forms at balanced ratio. Without shortcut saturation, gradient amplification does not begin, and the capable model fails for the same reason as the incapable one: the conditions for circuit reorganization do not arise.

Attention on the maximum-valued token tracks the same transition. Shortcut-trapped one-layer models put almost all of it there ($0.987 \pm 0.007$ at $r{=}0.9$, shortcut-consistent split) and little at $r{=}0.5$ ($0.114 \pm 0.109$), mirroring the QK Spearman $\rho$ of $0.987$ and $0.213$ in those cells. Within two-layer configurations it separates outcomes: at $r{=}0.9$ generalizing seeds end at $0.224 \pm 0.146$ (2L,~2H) and $0.150 \pm 0.095$ (2L,~1H) against $0.623 \pm 0.171$ and $0.544 \pm 0.221$ for the rest, with the same split at $r{=}0.7$ and none at $r{=}0.5$, where no shortcut circuit forms to displace. Displacement thus appears directly as attention withdrawn from the max-valued token (Appendix~\ref{app:maxattn}).

The position-shortcut task (first-element parity) exposes the same distinction through a complementary lens. Because the shortcut is positional rather than magnitude-based, QK Spearman $\rho$ is not the appropriate fingerprint: the shortcut circuit need not rank tokens by magnitude, only concentrate attention on position~0. We therefore use \emph{position-0 attention weight} ($p_0$) as the shortcut indicator; Table~\ref{tab:fesp_fingerprint} reports the results.

The 2L, 2H model at $r{=}0.9$ shows the same displacement signature as the magnitude-shortcut task: $p_0$ peaks at 0.671 during training and falls to 0.307 by the generalization epoch while OV parity Spearman reaches 0.813. One-layer models show a qualitatively different pattern---for the 1L, 1H model $p_0$ never peaks substantially (max 0.100 at $r{=}0.9$)---consistent with the positional shortcut being implemented via embeddings rather than attention concentration, the same distributed mechanism as the 1-layer magnitude shortcut. The ablation direction is consistent with this interpretation: zeroing attention in the shortcut-trapped 1L, 1H, $r{=}0.9$ model \emph{increases} adversarial accuracy by $0.42 \pm 0.16$ (suggesting attention implements a harmful shortcut), while the same ablation in the generalizing 2L, 2H model \emph{decreases} it by $0.28 \pm 0.30$ (suggesting attention now contributes to the robust solution).

\begin{table}[t]
  \caption{FESP positional shortcut fingerprint across training. $p_0$ (peak) is the maximum position-0 attention weight across all checkpoints. $p_0$ (at generalization) is measured at the first epoch at which adversarial accuracy reaches 100\% for generalizing models; $^{\dagger}$ marks the one configuration that never generalizes, where the final checkpoint is substituted. OV parity Spearman is reported at the final checkpoint (the at-generalization value is undefined for this positional-shortcut task). Entries without a standard deviation derive from a single generalizing seed.}
  \label{tab:fesp_fingerprint}
  \centering
  \small
  \begin{tabular}{llll}
    \toprule
    Config & $p_0$ (peak) & $p_0$ (at gen.) & OV parity (final) \\
    \midrule
    1L, 1H, $r{=}0.5$ (balanced)         & $0.015 \pm 0.025$ & $0.025$ & $-0.055 \pm 0.259$ \\
    1L, 1H, $r{=}0.9$ (shortcut-trapped) & $0.100 \pm 0.161$ & $0.098 \pm 0.015$ & $0.597 \pm 0.618$ \\
    2L, 2H, $r{=}0.5$ (balanced-trapped) & $0.152 \pm 0.163$ & $0.008 \pm 0.040$$^{\dagger}$ & $0.131 \pm 0.142$ \\
    2L, 2H, $r{=}0.9$ (generalized)      & $\mathbf{0.671 \pm 0.267}$ & $\mathbf{0.307 \pm 0.353}$ & $\mathbf{0.813 \pm 0.317}$ \\
    \bottomrule
  \end{tabular}
\end{table}

\subsection{Subset Trajectory: Two-Phase Accuracy Dynamics}

\begin{figure}[t]
  \centering
  \includegraphics[width=0.95\linewidth]{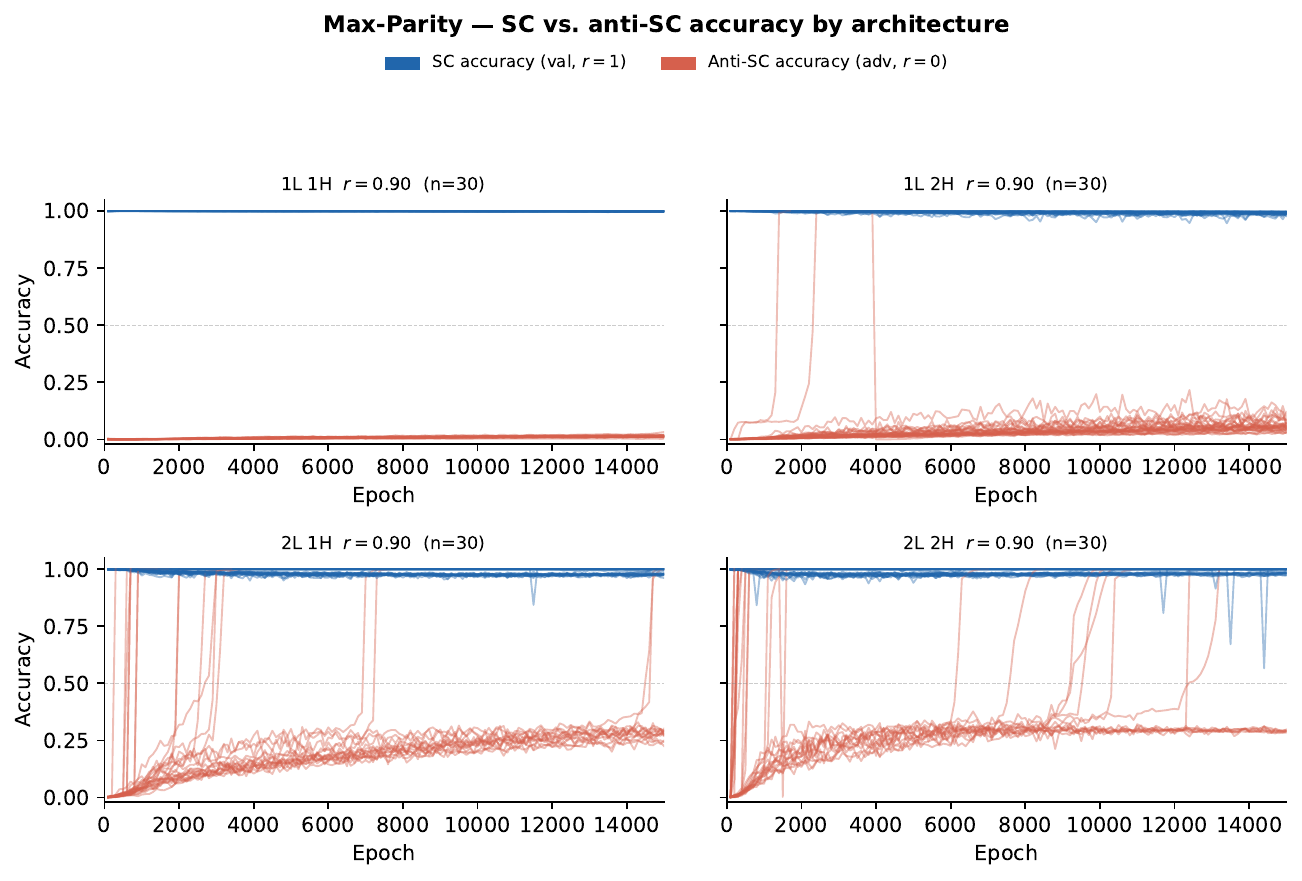}
  \caption{Per-epoch SC accuracy (blue, val split $r{=}1$) and anti-SC accuracy (red, adv split $r{=}0$) for all 30 seeds at $r{=}0.9$, weight decay $0.1$. Each anti-SC curve is drawn only up to the first epoch at which it reaches $100\%$, so a curve that ends at the top of a panel marks a seed that generalized; the post-generalization regression documented in Sections~\ref{sec:wd-transient} and~\ref{sec:circuit} is deliberately not shown here. \textbf{(a)}~1-layer, 1-head: all seeds permanently trapped in Phase~1 (highest anti-SC accuracy reached by any seed is $0.03$). \textbf{(b)}~1-layer, 2-head: similarly trapped. One seed reaches $100\%$ and holds it; a second rises to $99.95\%$ near epoch 2400 and then collapses back to $0.03$, which is the untruncated excursion visible in the panel. Under our definition only the first counts as generalizing. \textbf{(c)}~2-layer, 1-head: 16/30 seeds exhibit Phase~2 recovery. \textbf{(d)}~2-layer, 2-head: 23/30 seeds exhibit Phase~2 recovery, with seed-dependent onset.}
  \label{fig:subset_trajectory}
\end{figure}

Figure~\ref{fig:subset_trajectory} plots accuracy separately on the shortcut-consistent (SC, val split at $r{=}1$) and anti-shortcut (adv split at $r{=}0$) subsets for all 30 seeds at $r{=}0.9$, one panel per architecture. In both 1-layer models (panels a--b), SC accuracy rises to near-perfect within the first logged epoch while anti-shortcut accuracy stays near zero throughout---well below the 0.5 chance line, and training remains in Phase~1 for the full 15{,}000 epochs in all but one seed.

In 2-layer models (panels c--d), the same Phase~1 pattern appears initially: SC accuracy reaches near-perfect while anti-shortcut accuracy collapses toward zero. Phase~2 then begins at a seed-dependent epoch, ranging from a few hundred to over 10{,}000 epochs: anti-shortcut accuracy rises toward 1.0 while SC accuracy remains near-perfect. This trajectory is consistent with generalization being additive---shortcut-consistent performance is retained as adversarial robustness is acquired. A few seeds show brief SC dips during Phase~2, visible as downward spikes in panels (c--d); these recover, and mean final SC accuracy is $0.990$ (2L~1H) and $0.992$ (2L~2H). The Phase~2 onset is abrupt for some seeds and gradual for others, consistent with the wide seed-to-seed variance in generalization epochs. Head count modulates the fraction of seeds that enter Phase~2: 16/30 for 2L~1H (panel c) and 23/30 for 2L~2H (panel d).

\section{Discussion}
\label{sec:discussion}

\paragraph{Gradient amplification.}
Once the shortcut circuit saturates, its examples' gradients vanish while the misclassified anti-shortcut minority keeps producing large signals. Counting examples alone predicts a ratio of $r/(1-r)$, or $9{:}1$ at $r{=}0.9$. This sets a floor rather than an estimate: the 1-layer models, which stay on the shortcut, sit close to it ($8.8\times$ at 1H and $10.6\times$ at 2H), while the 2-layer models overshoot it substantially ($16.8\times$ at 2H and $26.7\times$ at 1H; the 1H mean is drawn up by a heavy right tail, and its median of $12.1\times$ still exceeds the floor). Example counts therefore explain why amplification appears at all, but not its size in the models that reorganize---there, the circuit reorganization itself appears to suppress the shortcut gradient further, beyond what the class proportions require.

\paragraph{Why capacity matters.}
The shortcut circuit occupies representational capacity. In a one-layer model the shortcut and true-rule circuits compete for the same parameters, and the shortcut---reinforced by 90\% of training examples---has the stronger attractor. In a two-layer model the second layer can encode the true rule while the first retains the shortcut, and the amplified adversarial gradient appears to route through the second layer. Loss-landscape sharpness for these runs (Appendix~\ref{app:sharpness}) does not by itself separate the two.

\paragraph{Why balanced data appears insufficient.}
At $r{=}0.5$ the shortcut reaches only 50\% training accuracy and gives no consistent signal. The two subsets contribute gradients of near-equal magnitude (ratio $0.99$ versus $16.8$ at $r{=}0.9$), so neither circuit consolidates even though the early conflict resolves: it is the absence of amplification, not unresolved conflict, that accompanies failure (Section~\ref{sec:gen-rate}).

\paragraph{Implications for training practice.}
Our findings are limited to small synthetic transformers, but they suggest the relationship between dataset balance, model capacity, and generalization is more complex than standard prescriptions assume.

\paragraph{Limitations.}
\begin{itemize}
  \item Despite 30 seeds per configuration for the four primary tasks (10 for the ablation variants), variance remains high for some settings (e.g., a generalization-epoch standard deviation of ${\sim}3{,}700$ for the 2L, 2H model at $r{=}0.9$, weight decay 0.4).
  \item Effect size varies across tasks (peak generalization 93\% for FE-Mod3 but 27\% for MM-Mod3), suggesting that the difficulty of the true computation relative to the shortcut modulates the effect. How this scales to real-world tasks remains open.
  \item The capacity threshold is the least portable of our findings. On MM-Mod3 the ordering inverts outright---a 1-layer model attains 50\%, above every 2-layer configuration---and on the easier n11 variant the threshold falls below two layers. The ratio effect replicates on all four tasks; the depth effect does not (Appendix~\ref{app:mod3_analysis}).
  \item We compare ratios and architectures, not debiasing algorithms. Our balanced conditions correspond to the target that majority subsampling moves toward, but we do not run group DRO, reweighting, or subsampling themselves, so we cannot rank our observation against those methods.
  \item The mechanistic timeline statistics (circuit formation epoch, conflict-resolution epoch) have high within-configuration variance, so seed-conditional analyses are more informative than per-configuration averages.
\end{itemize}

\section{Conclusion}

We find that data imbalance---conventionally viewed as a problem to correct---can promote robust generalization in capable models under spurious correlations, and on both binary tasks hurts it below a capacity threshold that lies between one and two layers. The dynamics are consistent with gradient amplification: as the shortcut saturates, the anti-shortcut minority becomes the dominant gradient source. The ternary-label experiments show that what matters is deviation of $r$ from the random-chance baseline: at the null ratio no two-layer configuration generalizes in more than 1 of 30 seeds, and imbalance in either direction restores generalization---fully on the ternary position-shortcut task (80--93\%) and only partially on the harder ternary magnitude-shortcut task (up to 27\%). The ratio effect replicates on all four tasks; the capacity effect does not, inverting on the ternary magnitude-shortcut task, and we take locating that boundary to be the natural next step.

\bibliographystyle{iclr2027_conference}
\bibliography{reference}

@article{power2022grokking,
  title={Grokking: Generalization beyond overfitting on small algorithmic datasets},
  author={Power, Alethea and Burda, Yuri and Edwards, Harri and Babuschkin, Igor and Misra, Vedant},
  journal={arXiv preprint arXiv:2201.02177},
  year={2022}
}

@inproceedings{nanda2023progress,
  title={Progress measures for grokking via mechanistic interpretability},
  author={Nanda, Neel and Chan, Lawrence and Lieberum, Tom and Smith, Jess and Steinhardt, Jacob},
  booktitle={International Conference on Learning Representations},
  year={2023}
}

@inproceedings{barak2022hidden,
  title={Hidden progress in deep learning: {SGD} learns parities near the computational limit},
  author={Barak, Boaz and Edelman, Benjamin L and Goel, Surbhi and Kakade, Sham M and Malach, Eran and Zhang, Cyril},
  booktitle={Advances in Neural Information Processing Systems},
  volume={35},
  year={2022}
}

@article{geirhos2020shortcut,
  title={Shortcut learning in deep neural networks},
  author={Geirhos, Robert and Jacobsen, J{\"o}rn-Henrik and Michaelis, Claudio and Zemel, Richard and Brendel, Wieland and Bethge, Matthias and Wichmann, Felix A},
  journal={Nature Machine Intelligence},
  volume={2},
  number={11},
  pages={665--673},
  year={2020}
}

@inproceedings{shah2020pitfalls,
  title={The pitfalls of simplicity bias in neural networks},
  author={Shah, Harshay and Tamuly, Kaustav and Raghunathan, Aditi and Jain, Prateek and Netrapalli, Praneeth},
  booktitle={Advances in Neural Information Processing Systems},
  volume={33},
  year={2020}
}

@inproceedings{sinha2020consistency,
  title={Class-Wise Difficulty-Balanced Loss for Solving Class-Imbalance},
  author={Sinha, Saptarshi and Ohashi, Hiroki and Nakamura, Katsuyuki},
  booktitle={Asian Conference on Computer Vision},
  pages={549--565},
  year={2020},
  organization={Springer}
}

@inproceedings{mansilla2021domain,
  title={Domain generalization via gradient surgery},
  author={Mansilla, Lucas and Echeveste, Rodrigo and Milone, Diego H and Ferrante, Enzo},
  booktitle={Proceedings of the IEEE/CVF International Conference on Computer Vision},
  pages={6630--6638},
  year={2021}
}

@inproceedings{teney2025simplicity,
  title={Do We Always Need the Simplicity Bias? Looking for Optimal Inductive Biases in the Wild},
  author={Teney, Damien and Jiang, Liangze and Gogianu, Florin and Abbasnejad, Ehsan},
  booktitle={Proceedings of the IEEE/CVF Conference on Computer Vision and Pattern Recognition},
  pages={79--90},
  year={2025}
}

@inproceedings{hermann2024foundations,
  title={On the Foundations of Shortcut Learning},
  author={Hermann, Katherine L. and Mobahi, Hossein and Fel, Thomas and Mozer, Michael C.},
  booktitle={The Twelfth International Conference on Learning Representations},
  year={2024}
}

@inproceedings{sagawa2020distributionally,
  title={Distributionally Robust Neural Networks for Group Shifts: On the Importance of Regularization for Worst-Case Generalization},
  author={Sagawa, Shiori and Koh, Pang Wei and Hashimoto, Tatsunori B and Liang, Percy},
  booktitle={International Conference on Learning Representations},
  year={2020}
}

@article{olsson2022context,
  title={In-context Learning and Induction Heads},
  author={Olsson, Catherine and Elhage, Nelson and Nanda, Neel and Joseph, Nicholas and DasSarma, Nova and Henighan, Tom and Mann, Ben and Askell, Amanda and Bai, Yuntao and Chen, Anna and Conerly, Tom and Drain, Dawn and Ganguli, Deep and Hatfield-Dodds, Zac and Hernandez, Danny and Johnston, Scott and Jones, Andy and Kernion, Jackson and Lovitt, Liane and Ndousse, Kamal and Amodei, Dario and Brown, Tom and Clark, Jack and Kaplan, Jared and McCandlish, Sam and Olah, Chris},
  journal={arXiv preprint arXiv:2209.11895},
  year={2022}
}

@article{you2025uncovering,
  title={Uncovering memorization effect in the presence of spurious correlations},
  author={You, Chenyu and Dai, Haocheng and Min, Yifei and Sekhon, Jasjeet S. and Joshi, Sarang and Duncan, James S.},
  journal={Nature Communications},
  volume={16},
  number={1},
  pages={5424},
  year={2025},
  month={July},
  doi={10.1038/s41467-025-61531-5}
}

@inproceedings{sagawa2020overparameterization,
  title={An investigation of why overparameterization exacerbates spurious correlations},
  author={Sagawa, Shiori and Raghunathan, Aditi and Koh, Pang Wei and Liang, Percy},
  booktitle={International Conference on Machine Learning},
  pages={8346--8356},
  year={2020},
  organization={PMLR}
}

@inproceedings{zhong2023clock,
  title={The clock and the pizza: Two stories in mechanistic explanation of neural networks},
  author={Zhong, Ziqian and Liu, Ziming and Tegmark, Max and Andreas, Jacob},
  booktitle={Advances in Neural Information Processing Systems},
  volume={36},
  year={2023}
}


\appendix

\section{Extended Results: Weight Decay = 0.4}
\label{app:wd04}

Table~\ref{tab:main} reports weight decay 0.1. Table~\ref{tab:wd04} reports the corresponding results for weight decay 0.4. The main imbalance $\times$ capacity interaction persists: 2-layer models improve with ratio while 1-layer models do not.

\begin{table}[htbp]
  \caption{Generalization rate by configuration (weight decay = 0.4).}
  \label{tab:wd04}
  \centering
  \begin{tabular}{ll@{\hskip 1.5em}ll}
    \toprule
    \multicolumn{2}{c}{1-layer models} & \multicolumn{2}{c}{2-layer models} \\
    \cmidrule(r){1-2}\cmidrule(l){3-4}
    Config & Gen.\ Rate & Config & Gen.\ Rate \\
    \midrule
    1H, $r{=}0.5$ & 0/30 (0\%)   & \textbf{1H, $r{=}0.5$} & \textbf{2/30 (7\%)} \\
    1H, $r{=}0.7$ & 0/30 (0\%)   & \textbf{1H, $r{=}0.7$} & \textbf{7/30 (23\%)} \\
    1H, $r{=}0.9$ & 0/30 (0\%)   & \textbf{1H, $r{=}0.9$} & \textbf{15/30 (50\%)} \\
    2H, $r{=}0.5$ & 4/30 (13\%)  & \textbf{2H, $r{=}0.5$} & \textbf{1/30 (3\%)}  \\
    2H, $r{=}0.7$ & 2/30 (7\%)   & \textbf{2H, $r{=}0.7$} & \textbf{17/30 (57\%)} \\
    2H, $r{=}0.9$ & 3/30 (10\%)  & \textbf{2H, $r{=}0.9$} & \textbf{21/30 (70\%)} \\
    \bottomrule
  \end{tabular}
\end{table}

The 2L, 2H model at $r{=}0.9$ retains a 70\% generalization rate under higher weight decay. As described in Section~\ref{sec:wd-transient}, 15 of those 21 seeds hit the threshold only transiently before regressing to near-shortcut performance; only 6 consolidated into stable generalizing solutions.

\section{First Equal Sum Parity}
\label{app:fesp_analysis}

Section~\ref{sec:cross-shortcut} summarizes the behavioral comparison between the magnitude-shortcut and position-shortcut binary tasks for 2-layer models; this appendix reports the full per-configuration results for all 12 configurations and provides additional mechanistic discussion. While the \tsk{max\us{}parity\us{}sum\us{}parity} (MPSP) task focuses on a magnitude-based shortcut, the \tsk{first\us{}equal\us{}sum\us{}parity} (FESP) task employs a purely position-based shortcut. In the position-shortcut task, the parity of the first element in the sequence ($seq[0]$) is correlated with the total sum parity in shortcut-consistent examples.

\subsection{Contrast in Task Difficulty}

\begin{table}[htbp]
  \caption{Generalization rate by configuration for the FESP task (weight decay = 0.1). Bold rows highlight 2-layer models. With 30 seeds, 2-layer models clearly dominate and generalization rises with ratio (2L, 2H reaches 77\% at $r{=}0.9$), while 1-layer models remain low---the same imbalance $\times$ capacity interaction seen in the magnitude-shortcut task.}
  \label{tab:first-equal-sum-parity-weight-decay-0.1}
  \centering
  \begin{tabular}{ll@{\hskip 1.5em}ll}
    \toprule
    \multicolumn{2}{c}{1-layer models} & \multicolumn{2}{c}{2-layer models} \\
    \cmidrule(r){1-2}\cmidrule(l){3-4}
    Config & Gen.\ Rate & Config & Gen.\ Rate \\
    \midrule
    1H, $r{=}0.5$ & 1/30 (3\%)   & \textbf{1H, $r{=}0.5$} & \textbf{0/30 (0\%)}  \\
    1H, $r{=}0.7$ & 7/30 (23\%)  & \textbf{1H, $r{=}0.7$} & \textbf{16/30 (53\%)} \\
    1H, $r{=}0.9$ & 2/30 (7\%)   & \textbf{1H, $r{=}0.9$} & \textbf{20/30 (67\%)} \\
    2H, $r{=}0.5$ & 1/30 (3\%)   & \textbf{2H, $r{=}0.5$} & \textbf{0/30 (0\%)}  \\
    2H, $r{=}0.7$ & 10/30 (33\%) & \textbf{2H, $r{=}0.7$} & \textbf{20/30 (67\%)} \\
    2H, $r{=}0.9$ & 1/30 (3\%)   & \textbf{2H, $r{=}0.9$} & \textbf{23/30 (77\%)} \\
    \bottomrule
  \end{tabular}
\end{table}

Raising weight decay to 0.4 preserves the same ordering across configurations (Table~\ref{tab:first-equal-sum-parity-weight-decay-0.4}).

\begin{table}[htbp]
  \caption{Generalization rate by configuration (weight decay = 0.4). Bold rows highlight the 2-layer models.}
  \label{tab:first-equal-sum-parity-weight-decay-0.4}
  \centering
  \begin{tabular}{ll@{\hskip 1.5em}ll}
    \toprule
    \multicolumn{2}{c}{1-layer models} & \multicolumn{2}{c}{2-layer models} \\
    \cmidrule(r){1-2}\cmidrule(l){3-4}
    Config & Gen.\ Rate & Config & Gen.\ Rate \\
    \midrule
    1H, $r{=}0.5$ & 0/30 (0\%)  & \textbf{1H, $r{=}0.5$} & \textbf{2/30 (7\%)} \\
    1H, $r{=}0.7$ & 0/30 (0\%)  & \textbf{1H, $r{=}0.7$} & \textbf{15/30 (50\%)} \\
    1H, $r{=}0.9$ & 0/30 (0\%)  & \textbf{1H, $r{=}0.9$} & \textbf{16/30 (53\%)} \\
    2H, $r{=}0.5$ & 0/30 (0\%)  & \textbf{2H, $r{=}0.5$} & \textbf{0/30 (0\%)}  \\
    2H, $r{=}0.7$ & 2/30 (7\%)  & \textbf{2H, $r{=}0.7$} & \textbf{22/30 (73\%)} \\
    2H, $r{=}0.9$ & 0/30 (0\%)  & \textbf{2H, $r{=}0.9$} & \textbf{20/30 (67\%)} \\
    \bottomrule
  \end{tabular}
\end{table}

As shown in Tables~\ref{tab:first-equal-sum-parity-weight-decay-0.1} and~\ref{tab:first-equal-sum-parity-weight-decay-0.4}, the position-shortcut task generalizes comparably to the magnitude-shortcut task once enough seeds are sampled. At high capacity and high imbalance ($r=0.9, \text{2L, 2H}$), the generalization rate reaches 77\%, matching the 77\% of the magnitude-shortcut task at the same setting. The two shortcut types are therefore dislodged with similar ease: a fixed positional pointer is no more resistant to the amplified adversarial gradient than a magnitude-based attention bias.

The two shortcuts are nonetheless processed differently by the Transformer, even though displacement ultimately proceeds at a comparable rate:
\begin{itemize}
    \item \textbf{Magnitude shortcut (MPSP):} Magnitude is often encoded in token embedding norms, so an attention head can implement this shortcut cheaply, and it saturates quickly at high $r$.
    \item \textbf{Position shortcut (FESP):} The positional shortcut requires absolute positional awareness. Moving beyond it requires both acquiring parity-sum representations and displacing the attention weight at $pos=0$. Despite this structural difference, the saturated positional circuit is displaced at a rate comparable to the magnitude circuit.
\end{itemize}

\subsection{The 1-Layer Capacity Threshold and Shortcut-Trapping}

A critical finding in our experiments is that the beneficial effect of data imbalance is conditioned on model capacity. For the position-shortcut task, 1-layer models exhibit a reverse trend compared to their 2-layer counterparts: generalization peaks at the intermediate ratio $r{=}0.7$ and collapses to near-zero at $r{=}0.9$, the opposite of the 2-layer trend where generalization rises with ratio.

\paragraph{Pattern in 1-Layer Models.}
In 1-layer models, the pattern inverts. One interpretation is that the architecture lacks the capacity to maintain competing circuits. At a balanced ratio ($r=0.5$), the shortcut provides no consistent signal, and parameters are not strongly drawn toward either feature. At $r=0.9$, the shortcut is so statistically dominant and computationally cheap---requiring only a simple positional pointer to $pos=0$---that the limited parameters appear to be entirely consumed by the shortcut circuit. 

\paragraph{Adversarial Accuracy as a Measure of Capture.} 
The near-zero adversarial accuracy observed in 1-layer, high-ratio models (e.g., 0.044 for 1L, 2H at $r=0.9$) is the clearest evidence of this ``shortcut-trapping''. Because the adversarial set is constructed so that the shortcut contradicts the true label, a model that has perfectly learned the shortcut will consistently predict the wrong parity, resulting in near-0\% accuracy. This is consistent with the following interpretation.
\begin{itemize}
    \item There is a capacity threshold---in this case, between one and two transformer layers---below which imbalance is harmful because training converges onto the easiest feature.
    \item Above this threshold, the additional parameters appear to allow the amplified adversarial gradient signal to reorganize representations without fully displacing initial shortcut performance.
\end{itemize}

\section{n11 Variant Replication}
\label{app:n11}

The \tsk{max\us{}parity\us{}sum\us{}parity\us{}n11} variant uses integers drawn from $[0, 11)$ instead of $[0, 20)$, producing a lower-variance max-element distribution. This makes the sum parity task structurally easier (fewer distinct values, shorter effective range). Tables~\ref{tab:n11} and~\ref{tab:n11_wd04} report generalization rates for weight decay 0.1 and 0.4 respectively; all per-seed counts are out of 10 seeds.

\begin{table}[htbp]
  \caption{n11 variant: generalization rate (weight decay = 0.1).}
  \label{tab:n11}
  \centering
  \begin{tabular}{ll@{\hskip 1.5em}ll}
    \toprule
    \multicolumn{2}{c}{1-layer models} & \multicolumn{2}{c}{2-layer models} \\
    \cmidrule(r){1-2}\cmidrule(l){3-4}
    Config & Gen.\ Rate & Config & Gen.\ Rate \\
    \midrule
    1H, $r{=}0.5$ & 5/10 (50\%)  & \textbf{1H, $r{=}0.5$} & \textbf{7/10 (70\%)}   \\
    1H, $r{=}0.7$ & 8/10 (80\%)  & \textbf{1H, $r{=}0.7$} & \textbf{9/10 (90\%)}   \\
    1H, $r{=}0.9$ & 1/10 (10\%)  & \textbf{1H, $r{=}0.9$} & \textbf{10/10 (100\%)} \\
    2H, $r{=}0.5$ & 7/10 (70\%)  & \textbf{2H, $r{=}0.5$} & \textbf{10/10 (100\%)} \\
    2H, $r{=}0.7$ & 7/10 (70\%)  & \textbf{2H, $r{=}0.7$} & \textbf{10/10 (100\%)} \\
    2H, $r{=}0.9$ & 4/10 (40\%)  & \textbf{2H, $r{=}0.9$} & \textbf{10/10 (100\%)} \\
    \bottomrule
  \end{tabular}
\end{table}

The same configurations under weight decay 0.4 are reported in Table~\ref{tab:n11_wd04}.

\begin{table}[htbp]
  \caption{n11 variant: generalization rate (weight decay = 0.4).}
  \label{tab:n11_wd04}
  \centering
  \begin{tabular}{ll@{\hskip 1.5em}ll}
    \toprule
    \multicolumn{2}{c}{1-layer models} & \multicolumn{2}{c}{2-layer models} \\
    \cmidrule(r){1-2}\cmidrule(l){3-4}
    Config & Gen.\ Rate & Config & Gen.\ Rate \\
    \midrule
    1H, $r{=}0.5$ & 0/10 (0\%)   & \textbf{1H, $r{=}0.5$} & \textbf{8/10 (80\%)}   \\
    1H, $r{=}0.7$ & 0/10 (0\%)   & \textbf{1H, $r{=}0.7$} & \textbf{6/10 (60\%)}   \\
    1H, $r{=}0.9$ & 0/10 (0\%)   & \textbf{1H, $r{=}0.9$} & \textbf{8/10 (80\%)}   \\
    2H, $r{=}0.5$ & 5/10 (50\%)  & \textbf{2H, $r{=}0.5$} & \textbf{10/10 (100\%)} \\
    2H, $r{=}0.7$ & 6/10 (60\%)  & \textbf{2H, $r{=}0.7$} & \textbf{8/10 (80\%)}   \\
    2H, $r{=}0.9$ & 0/10 (0\%)   & \textbf{2H, $r{=}0.9$} & \textbf{8/10 (80\%)}   \\
    \bottomrule
  \end{tabular}
\end{table}

Two findings replicate in the n11 variant. First, the capacity threshold effect persists: the 1L, 1H model is trapped by high imbalance ($r{=}0.9$, 10\% generalization rate) despite strong generalization at $r{=}0.7$ (80\%), mirroring the base task's 1-layer shortcut trap. Second, 2-layer models generalize robustly at all ratios, indicating that the imbalance $\times$ capacity interaction is not an artifact of the base task's specific integer range. With 10 seeds per configuration this variant is a replication check rather than a second primary result.

The n11 task sits at a lower overall difficulty: the 2L, 2H model achieves 100\% generalization even at balanced ratios ($r{=}0.5$), which does not occur in the base task. This suggests the capacity threshold for n11 falls below the 2L, 2H architecture. The location of the threshold thus appears to depend on the difficulty of the true rule relative to the shortcut---consistent with the interpretation in Section~\ref{sec:discussion}.

\section{Mod~3 Label Tasks}
\label{app:mod3_analysis}
The two mod~3 tasks use ternary labels (3 classes) instead of binary, so the random-chance shortcut correlation is $\nicefrac{1}{3}$ rather than $\nicefrac{1}{2}$. The ratio sweep is $r \in \{0.15, 0.33, 0.50\}$, where $r{=}0.33 \approx \nicefrac{1}{3}$ is the null point and $r{=}0.15$ and $r{=}0.50$ are imbalanced in opposite directions.
\subsection{First-Equal-Sum-Mod3 (FE-Mod3): Positional Shortcut, Ternary Label}
\begin{table}[htbp]
  \caption{Generalization rates for the FE-Mod3 task (weight decay = 0.1). At the null ratio $r{=}0.33$ (random-chance shortcut correlation), \textbf{no 2-layer model generalizes}. Both $r{=}0.15$ (anti-biased) and $r{=}0.50$ (biased) achieve 80--93\% generalization in 2-layer models, indicating that the relevant quantity is deviation from the random-chance baseline, not the direction of the imbalance.}
  \label{tab:femod3}
  \centering
  \begin{tabular}{ll@{\hskip 1.5em}ll}
    \toprule
    \multicolumn{2}{c}{1-layer models} & \multicolumn{2}{c}{2-layer models} \\
    \cmidrule(r){1-2}\cmidrule(l){3-4}
    Config & Gen.\ Rate & Config & Gen.\ Rate \\
    \midrule
    1H, $r{=}0.15$ & 7/30 (23\%)  & \textbf{1H, $r{=}0.15$} & \textbf{25/30 (83\%)} \\
    1H, $r{=}0.33$ & 11/30 (37\%) & \textbf{1H, $r{=}0.33$} & \textbf{0/30 (0\%)}   \\
    1H, $r{=}0.50$ & 11/30 (37\%) & \textbf{1H, $r{=}0.50$} & \textbf{24/30 (80\%)} \\
    2H, $r{=}0.15$ & 9/30 (30\%)  & \textbf{2H, $r{=}0.15$} & \textbf{28/30 (93\%)} \\
    2H, $r{=}0.33$ & 0/30 (0\%)   & \textbf{2H, $r{=}0.33$} & \textbf{0/30 (0\%)}   \\
    2H, $r{=}0.50$ & 2/30 (7\%)   & \textbf{2H, $r{=}0.50$} & \textbf{27/30 (90\%)} \\
    \bottomrule
  \end{tabular}
\end{table}

The corresponding results at weight decay 0.4 are given in Table~\ref{tab:femod3_wd04}.

\begin{table}[htbp]
  \caption{Generalization rates for the FE-Mod3 task (weight decay = 0.4). The U-shaped pattern in 2-layer models persists under higher weight decay, though peak generalization rates are attenuated.}
  \label{tab:femod3_wd04}
  \centering
  \begin{tabular}{ll@{\hskip 1.5em}ll}
    \toprule
    \multicolumn{2}{c}{1-layer models} & \multicolumn{2}{c}{2-layer models} \\
    \cmidrule(r){1-2}\cmidrule(l){3-4}
    Config & Gen.\ Rate & Config & Gen.\ Rate \\
    \midrule
    1H, $r{=}0.15$ & 1/30 (3\%)  & \textbf{1H, $r{=}0.15$} & \textbf{11/30 (37\%)} \\
    1H, $r{=}0.33$ & 3/30 (10\%) & \textbf{1H, $r{=}0.33$} & \textbf{0/30 (0\%)}   \\
    1H, $r{=}0.50$ & 9/30 (30\%) & \textbf{1H, $r{=}0.50$} & \textbf{12/30 (40\%)} \\
    2H, $r{=}0.15$ & 3/30 (10\%) & \textbf{2H, $r{=}0.15$} & \textbf{17/30 (57\%)} \\
    2H, $r{=}0.33$ & 0/30 (0\%)  & \textbf{2H, $r{=}0.33$} & \textbf{0/30 (0\%)}   \\
    2H, $r{=}0.50$ & 5/30 (17\%) & \textbf{2H, $r{=}0.50$} & \textbf{18/30 (60\%)} \\
    \bottomrule
  \end{tabular}
\end{table}

Table~\ref{tab:femod3} reveals a pronounced U-shaped dependence in 2-layer models. At $r{=}0.33$ (the ternary random-chance baseline), the shortcut is statistically uninformative: gradient signals from all three label classes are roughly balanced, and 0\% of 2-layer seeds generalize. At both $r{=}0.15$ (where 85\% of training examples are anti-shortcut) and $r{=}0.50$ (where 50\% are shortcut-consistent), the imbalance is associated with a gradient asymmetry and 80--93\% generalization. One-layer models show no such U-shape: the 1L, 1H configuration generalizes at similar rates across all three ratios, including the null (23--37\%), consistent with the capacity requirement of Section~\ref{sec:discussion}. This holds regardless of the \emph{direction} of imbalance---shortcut-biased or anti-shortcut-biased---consistent with the hypothesis that the relevant mechanism is shortcut saturation (or anti-shortcut saturation), not simply the direction of the imbalance.

The result suggests a unifying principle across all four tasks: what matters appears to be the distance of $r$ from the random-chance baseline ($\nicefrac{1}{2}$ for binary tasks, $\nicefrac{1}{3}$ for ternary tasks), not the absolute value of $r$. At the baseline, gradient conflict persists; away from it in either direction, one side appears to eventually saturate and its gradient collapses, amplifying the other.
\subsection{Max-Mod3-Sum-Mod3 (MM-Mod3): Magnitude Shortcut, Ternary Label}
\begin{table}[htbp]
  \caption{Generalization rates for the MM-Mod3 task (weight decay = 0.1). The effect is present but weaker than the ternary position-shortcut and binary magnitude-shortcut tasks.}
  \label{tab:mmmod3}
  \centering
  \begin{tabular}{ll@{\hskip 1.5em}ll}
    \toprule
    \multicolumn{2}{c}{1-layer models} & \multicolumn{2}{c}{2-layer models} \\
    \cmidrule(r){1-2}\cmidrule(l){3-4}
    Config & Gen.\ Rate & Config & Gen.\ Rate \\
    \midrule
    1H, $r{=}0.15$ & 8/30 (27\%)  & \textbf{1H, $r{=}0.15$} & \textbf{3/30 (10\%)}  \\
    1H, $r{=}0.33$ & 2/30 (7\%)   & \textbf{1H, $r{=}0.33$} & \textbf{0/30 (0\%)}   \\
    1H, $r{=}0.50$ & 15/30 (50\%) & \textbf{1H, $r{=}0.50$} & \textbf{0/30 (0\%)}   \\
    2H, $r{=}0.15$ & 0/30 (0\%)   & \textbf{2H, $r{=}0.15$} & \textbf{8/30 (27\%)}  \\
    2H, $r{=}0.33$ & 0/30 (0\%)   & \textbf{2H, $r{=}0.33$} & \textbf{1/30 (3\%)}   \\
    2H, $r{=}0.50$ & 2/30 (7\%)   & \textbf{2H, $r{=}0.50$} & \textbf{5/30 (17\%)} \\
    \bottomrule
  \end{tabular}
\end{table}

The corresponding results at weight decay 0.4 are given in Table~\ref{tab:mmmod3_wd04}.

\begin{table}[htbp]
  \caption{Generalization rates for the MM-Mod3 task (weight decay = 0.4). The effect remains weak; the $r{=}0.33$ null continues to suppress generalization.}
  \label{tab:mmmod3_wd04}
  \centering
  \begin{tabular}{ll@{\hskip 1.5em}ll}
    \toprule
    \multicolumn{2}{c}{1-layer models} & \multicolumn{2}{c}{2-layer models} \\
    \cmidrule(r){1-2}\cmidrule(l){3-4}
    Config & Gen.\ Rate & Config & Gen.\ Rate \\
    \midrule
    1H, $r{=}0.15$ & 0/30 (0\%)   & \textbf{1H, $r{=}0.15$} & \textbf{0/30 (0\%)}  \\
    1H, $r{=}0.33$ & 1/30 (3\%)   & \textbf{1H, $r{=}0.33$} & \textbf{0/30 (0\%)}  \\
    1H, $r{=}0.50$ & 11/30 (37\%) & \textbf{1H, $r{=}0.50$} & \textbf{0/30 (0\%)}  \\
    2H, $r{=}0.15$ & 0/30 (0\%)   & \textbf{2H, $r{=}0.15$} & \textbf{0/30 (0\%)}  \\
    2H, $r{=}0.33$ & 0/30 (0\%)   & \textbf{2H, $r{=}0.33$} & \textbf{0/30 (0\%)}  \\
    2H, $r{=}0.50$ & 1/30 (3\%)   & \textbf{2H, $r{=}0.50$} & \textbf{1/30 (3\%)} \\
    \bottomrule
  \end{tabular}
\end{table}

The magnitude-shortcut ternary task (MM-Mod3) shows a qualitatively similar but substantially weaker pattern (Tables~\ref{tab:mmmod3} and~\ref{tab:mmmod3_wd04}). The $r{=}0.33$ null continues to suppress generalization in 2-layer models. However, the imbalanced ratios yield only 17--27\% generalization for 2L, 2H---far below the 90--93\% reached in the ternary position-shortcut task and the 77\% of the binary magnitude-shortcut task. Two factors likely contribute. First, computing sum mod~3 requires distinguishing three residue classes across five elements, a harder aggregation than sum parity. Second, the magnitude shortcut for a ternary label (max element mod~3) does not rank tokens as unambiguously as the binary case, producing a noisier shortcut circuit that saturates less cleanly.

\paragraph{The capacity ordering inverts on this task.} MM-Mod3 is the one task in our set where depth does not help, and we state this plainly because it bounds the claim in Section~\ref{sec:gen-rate}. The 1L,~1H model at $r{=}0.50$ generalizes in 15/30 seeds (50\%)---the highest rate anywhere in Table~\ref{tab:mmmod3}, and above \emph{every} 2-layer cell, the best of which reaches 8/30 (27\%). The 2L,~1H model never generalizes at either imbalanced ratio (0/30 at both $r{=}0.33$ and $r{=}0.50$) while its 1-layer counterpart reaches 50\%. The same ordering holds at weight decay 0.4 (Table~\ref{tab:mmmod3_wd04}), where 1L,~1H at $r{=}0.50$ reaches 11/30 and no 2-layer cell exceeds 1/30, so this is not a single-hyperparameter artifact.

What survives on this task is the ratio effect rather than the capacity effect: the 1L,~1H row is $27\%$, $7\%$, $50\%$ across $r \in \{0.15, 0.33, 0.50\}$, still dipping at the null. We do not have a mechanistic account of why depth reverses here, and we flag it as the clearest limit on the capacity threshold reported in Section~\ref{sec:gen-rate}.

\section{Loss Landscape Sharpness}
\label{app:sharpness}

\begin{table}[htbp]
  \caption{Loss landscape sharpness at generalization onset and at training end. Sharpness at generalization onset is defined only for seeds that generalize; the final column averages all 30 seeds, so the two columns are not like-for-like (see text for the paired comparison). High final sharpness is not exclusive to generalizing models: the 2L,~2H model at $r{=}0.5$ (0\% generalization rate) reaches the highest final sharpness of any configuration shown.}
  \label{tab:sharpness}
  \centering
  \begin{tabular}{lll}
    \toprule
    Config & Sharpness at gen.\ onset & Sharpness (final) \\
    \midrule
    1L, 2H, $r{=}0.5$ (generalized) & $1.78 \pm 1.90$ & $25.28 \pm 10.08$ \\
    2L, 1H, $r{=}0.5$ & $3.92 \pm 2.84$ & $94.48 \pm 34.19$ \\
    2L, 1H, $r{=}0.9$ & $4.16 \pm 3.87$ & $41.97 \pm 49.44$ \\
    2L, 2H, $r{=}0.5$ (balanced-trapped) & --- & $95.25 \pm 14.01$ \\
    2L, 2H, $r{=}0.9$ & $8.05 \pm 13.47$ & $38.90 \pm 45.20$ \\
    \bottomrule
  \end{tabular}
\end{table}

The two columns of Table~\ref{tab:sharpness} are computed over different seed populations: sharpness at generalization onset exists only for seeds that generalize, while final sharpness averages all 30 seeds. Compared like for like, on the generalizing seeds alone, the apparent flat-to-sharp transition holds in only one of the four configurations, taking a majority of seeds increasing as the criterion. Median sharpness moves from $4.72$ to $16.33$ for 2L,~2H at $r{=}0.9$ (16/23 seeds increase), but from $0.79$ to $0.00$ for 1L,~2H at $r{=}0.5$ (0/3 increase), from $3.67$ to $0.31$ for 2L,~1H at $r{=}0.9$ (3/16 increase), and from $2.67$ to $5.42$ for 2L,~1H at $r{=}0.5$ (1/3 increase, a median that rises on a minority of seeds and on a cell too small to read). Sharpness is likewise not a predictor of generalization across configurations: 2L,~2H at $r{=}0.5$, with a 0\% generalization rate, reaches the highest final sharpness of any configuration ($95.25 \pm 14.01$). We therefore make no sharpness claim beyond describing the solutions found.

\clearpage
\section{Attention to the Max-Valued Token}
\label{app:maxattn}

Section~\ref{sec:qkov} reports that shortcut displacement is visible as attention withdrawn from the maximum-valued token. Table~\ref{tab:maxattn} gives the per-configuration values behind that statement, at the final checkpoint and weight decay $0.1$; $p_\text{max}$ is defined in Appendix~\ref{app:methods}.

\begin{table}[htbp]
  \caption{Max-token attention $p_\text{max}$ at the final checkpoint (wd = 0.1). The first two columns average all 30 seeds on each split; the last two split the shortcut-consistent column by outcome, with seed counts in parentheses. Two patterns stand out. One-layer models at $r \geq 0.7$ end with nearly all attention on the max-valued token, rising to $0.987$ at $r{=}0.9$ where no seed generalizes. Wherever a configuration contains both outcomes and a shortcut circuit has formed ($r \geq 0.7$), generalizing seeds sit far below non-generalizing ones. At $r{=}0.5$, where the QK fingerprint shows no shortcut circuit forms (Table~\ref{tab:qkov}), both groups sit low and the separation vanishes.}
  \label{tab:maxattn}
  \centering
  \small
  \begin{tabular}{ll@{\hskip 1em}ll@{\hskip 1.5em}ll}
    \toprule
    & & \multicolumn{2}{c}{All seeds} & \multicolumn{2}{c}{Shortcut-consistent split, by outcome} \\
    \cmidrule(r){3-4}\cmidrule(l){5-6}
    Config & $r$ & SC split & Adv split & Generalizing & Non-generalizing \\
    \midrule
    1L, 1H & 0.5 & $0.114 \pm 0.109$ & $0.180 \pm 0.134$ & $0.199 \pm 0.132$ {\scriptsize (10)} & $0.072 \pm 0.066$ {\scriptsize (20)} \\
    1L, 1H & 0.7 & $0.886 \pm 0.161$ & $0.951 \pm 0.177$ & $0.308 \pm 0.034$ {\scriptsize (2)} & $0.927 \pm 0.035$ {\scriptsize (28)} \\
    1L, 1H & 0.9 & $\mathbf{0.987 \pm 0.007}$ & $1.000 \pm 0.000$ & --- {\scriptsize (0)} & $0.987 \pm 0.007$ {\scriptsize (30)} \\
    1L, 2H & 0.5 & $0.102 \pm 0.160$ & $0.195 \pm 0.234$ & $0.140 \pm 0.038$ {\scriptsize (3)} & $0.098 \pm 0.168$ {\scriptsize (27)} \\
    1L, 2H & 0.7 & $0.624 \pm 0.245$ & $0.678 \pm 0.281$ & $0.147 \pm 0.028$ {\scriptsize (3)} & $0.677 \pm 0.195$ {\scriptsize (27)} \\
    1L, 2H & 0.9 & $0.754 \pm 0.212$ & $0.821 \pm 0.224$ & $0.134$ {\scriptsize (1)} & $0.776 \pm 0.180$ {\scriptsize (29)} \\
    2L, 1H & 0.5 & $0.113 \pm 0.148$ & $0.120 \pm 0.141$ & $0.108 \pm 0.035$ {\scriptsize (3)} & $0.113 \pm 0.156$ {\scriptsize (27)} \\
    2L, 1H & 0.7 & $0.485 \pm 0.270$ & $0.542 \pm 0.279$ & $0.178 \pm 0.132$ {\scriptsize (8)} & $0.597 \pm 0.215$ {\scriptsize (22)} \\
    2L, 1H & 0.9 & $0.334 \pm 0.258$ & $0.370 \pm 0.272$ & $\mathbf{0.150 \pm 0.095}$ {\scriptsize (16)} & $\mathbf{0.544 \pm 0.221}$ {\scriptsize (14)} \\
    2L, 2H & 0.5 & $0.078 \pm 0.103$ & $0.114 \pm 0.114$ & --- {\scriptsize (0)} & $0.078 \pm 0.103$ {\scriptsize (30)} \\
    2L, 2H & 0.7 & $0.384 \pm 0.255$ & $0.387 \pm 0.255$ & $0.141 \pm 0.109$ {\scriptsize (13)} & $0.570 \pm 0.155$ {\scriptsize (17)} \\
    2L, 2H & 0.9 & $0.317 \pm 0.227$ & $0.326 \pm 0.247$ & $\mathbf{0.224 \pm 0.146}$ {\scriptsize (23)} & $\mathbf{0.623 \pm 0.171}$ {\scriptsize (7)} \\
    \bottomrule
  \end{tabular}
\end{table}

Two caveats bound what this measurement supports. It is an association, not an intervention: we do not force $p_\text{max}$ down and observe generalization. And it is measured at the final checkpoint, so for the seeds that regress after generalizing (Section~\ref{sec:circuit}) it describes where training ended rather than the state at the generalization epoch.

\section{Analysis Methodology}
\label{app:methods}

\paragraph{Circuit identification.}
Let $s_\text{val}$ and $s_\text{adv}$ be the change in accuracy on the shortcut-consistent and adversarial splits when a head's output is zeroed (positive = head helps that split), and let $g_\text{fwd}$, $g_\text{rev}$ be the activation-patching generalization scores in the val$\to$adv and adv$\to$val directions respectively. A head carries the \texttt{shared} label at checkpoint $t$ when all four scores are positive: $s_\text{val} > 0$ and $s_\text{adv} > 0$ (it contributes to both splits) and $g_\text{fwd} > 0$ and $g_\text{rev} > 0$ (cross-split patching transfers in both directions). Split selectivity, $|s_\text{val} - s_\text{adv}|\,/\,(|s_\text{val}| + |s_\text{adv}|)$, separates the two \emph{shortcut} labels from each other and does not gate the \texttt{shared} label. Two metrics are built on this label, and they differ in one respect, which we state because it affects how they should be compared. The \emph{epoch of first robust head} is the earliest checkpoint at which any attention head across all layers is labelled \texttt{shared} \emph{and} has mean patching score $\bar{g} = (g_\text{fwd} + g_\text{rev})/2 > 0.5$; this additional threshold keeps weakly transferring heads from triggering the event. The \emph{robust circuit fraction at generalization} is the share of all attention heads labelled \texttt{shared} --- without the $\bar{g}$ threshold --- at the last checkpoint at or before the generalization epoch, i.e.\ the first epoch where adversarial accuracy reaches 100\%. The \emph{final robust circuit fraction} is the corresponding share at the last training checkpoint; it is lower than the at-generalization fraction in runs that undergo post-generalization slingshot regression.

\paragraph{Max-valued-token attention.}
At each logged checkpoint we record the attention the final position places on every input position, for one fixed probe sequence per split, and average over all heads and layers. The \emph{max-token attention} $p_\text{max}$ is the weight on the position holding the largest integer in that sequence. Integer positions are identified by token value; the \texttt{START} and \texttt{END} tokens carry the two values above the integer range and are excluded, since \texttt{START} would otherwise be the largest value in every sequence. Because a single probe sequence is used, $p_\text{max}$ reports what one representative input elicits rather than a dataset average; its seed-to-seed standard deviation is nonetheless small in the trapped configurations ($0.007$ for 1L,~1H at $r{=}0.9$), and the probe is identical across all seeds and configurations, so the comparisons in Table~\ref{tab:maxattn} are like-for-like.

\paragraph{Gradient attribution.}
At each logged checkpoint, gradients are computed separately on the shortcut-consistent and adversarial splits using the same cross-entropy loss. Gradient cosine similarity is the cosine of the angle between the two flattened gradient vectors. The gradient ratio is the adversarial gradient norm divided by the shortcut-consistent gradient norm. The conflict-resolution epoch is the first epoch at which cosine similarity exceeds $-0.1$, i.e.\ the first epoch at which the two gradients are no longer strongly opposed.

\paragraph{QK/OV circuit analysis.}
For each attention head, the QK circuit is characterized by computing Spearman $\rho$ between query-key dot products and the magnitude rank of elements at each position. A value near 1 indicates the head preferentially attends to higher-valued tokens (magnitude-sorting shortcut circuit). The OV parity Spearman is computed by projecting the output-value matrix onto a parity probe direction trained on the shortcut-consistent split, then measuring rank correlation between projected values and ground-truth element parities. A high value indicates the output matrix encodes parity information. Ablation drops are measured by zeroing the head's output and recording the resulting change in accuracy.

\section{Computational Resources}
\label{app:compute}
All experiments were conducted using TPU v5-e8 accelerators provided via Kaggle. A single experimental configuration required approximately 7 hours of wall-clock time per 10 seeds to complete the 15,000-epoch training schedule; the two primary binary tasks and the two mod~3 tasks were each extended to 30 seeds per configuration.



\end{document}